\documentclass[letterpaper]{article} 
\usepackage{aaai20}  
\usepackage{times}  
\usepackage{helvet} 
\usepackage{courier}  
\usepackage[hyphens]{url}  
\usepackage{graphicx} 
\usepackage{graphicx}  
\usepackage{enumitem}
\usepackage{latexsym}
\usepackage{multirow}
\usepackage{bm}
\usepackage{amsmath,amssymb} 

\usepackage[ruled]{algorithm}
\usepackage{algorithmic}
\usepackage{multirow}
\usepackage{diagbox}

\usepackage{subcaption}

\begin{document}
%
\title{Masking Orchestration: Multi-task Pretraining \\for Multi-role Dialogue Representation Learning}
\author{Tianyi Wang*\textsuperscript{1},
Yating Zhang*\textsuperscript{1},
Xiaozhong Liu\textsuperscript{2},
Changlong Sun\textsuperscript{1},
Qiong Zhang\textsuperscript{1}\\
\thanks{Both authors contributed equally to this research.}
\textsuperscript{1}{Alibaba Group, Hangzhou, Zhejiang, China}\\
\textsuperscript{2}{Indiana University Bloomington, Bloomington, Indiana, USA}\\
will.wty@alibaba-inc.com,
ranran.zyt@alibaba-inc.com\\
liu237@indiana.edu,
changlong.scl@taobao.com,
qz.zhang@alibaba-inc.com}

\maketitle
\begin{abstract}
\begin{quote}
Multi-role dialogue understanding comprises a wide range of diverse tasks such as question answering, act classification, dialogue summarization etc. While dialogue corpora are abundantly available, labeled data, for specific learning tasks, can be highly scarce and expensive. In this work, we investigate dialogue context representation learning with various types unsupervised pretraining tasks where the training objectives are given naturally according to the nature of the utterance and the structure of the multi-role conversation. Meanwhile, in order to locate essential information for dialogue summarization/extraction, the pretraining process enables external knowledge integration. The proposed fine-tuned pretraining mechanism is comprehensively evaluated via three different dialogue datasets along with a number of downstream dialogue-mining tasks. Result shows that the proposed pretraining mechanism significantly contributes to all the downstream tasks without discrimination to different encoders.

\end{quote}
\end{abstract}
\section{Introduction}
Multi-role dialogue mining is a novel topic of critical importance, and it offers powerful potentials for a number of scenarios, e.g, the court debate in civil trial where parties from different camps (plaintiff, defendant, witness, judge etc.) are actively involved, the customer service calls arisen from agent(s) and customer, the business meeting engaged with multi-members. Unfortunately, compared with classical textual data, the labeled multi-role dialogue corpus is scarce and expensive. Unsupervised learning, as a critical alternative, can alleviate this problem, while, based on prior experience\cite{mikolov2013efficient,devlin2018bert,radford2018improving,peters2018deep}, pretraining for complex text data can provide an enhanced content representation for the downstream tasks. In this study, we investigate an innovative problem, multi-role dialogue pretraining for various kinds of NLP tasks.
\begin{figure}[!t]
    \centering
    \includegraphics[width=\linewidth]{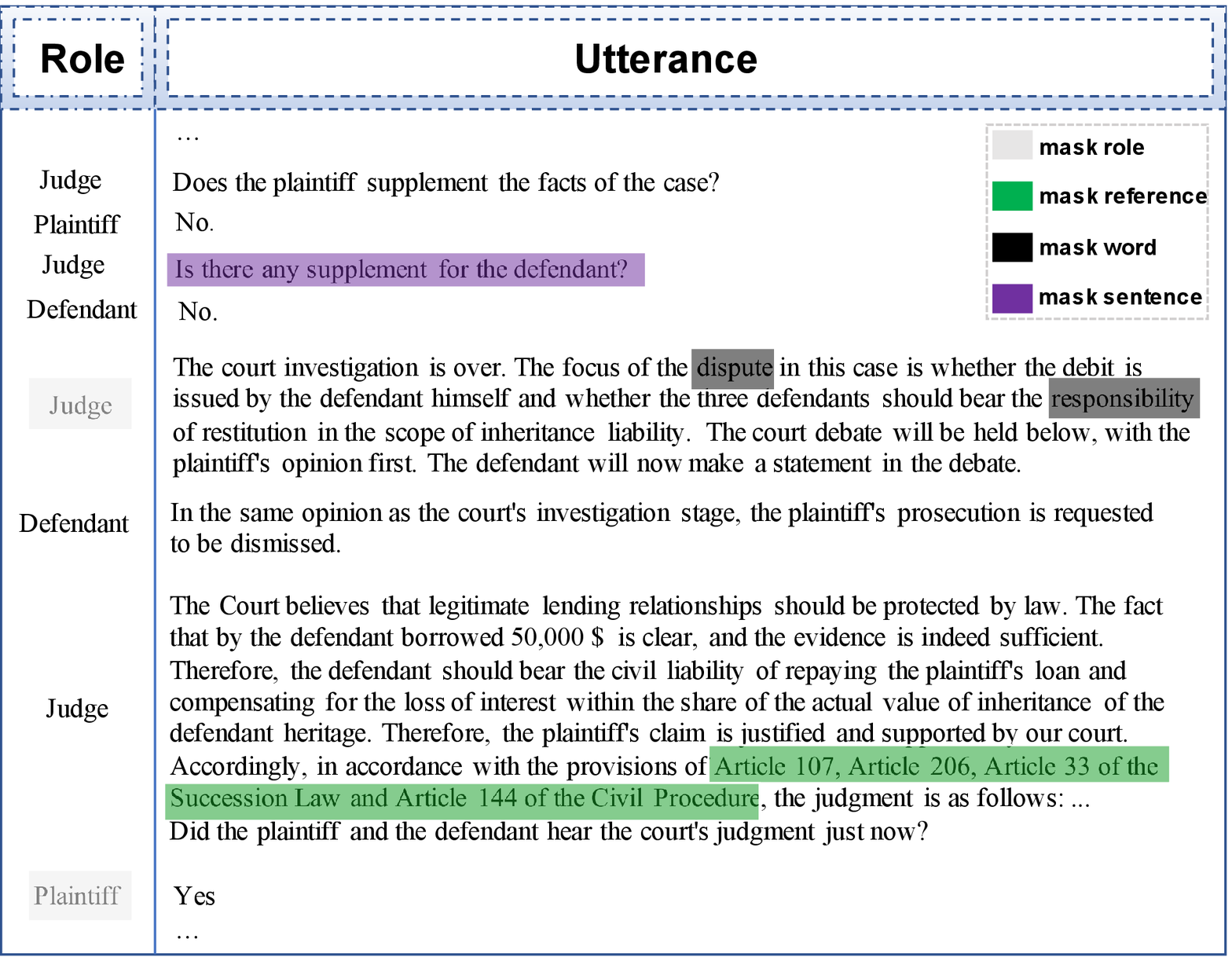}
    \caption{Example Dialogue in Court Debate Dataset \label{fig:case}}
    \vspace{-18pt}
\end{figure}

Indubitably, multi-role dialogue is more complex in its discourse structure and sometimes implicit/ambiguous in its semantics. Two major challenges should be highlighted for topic. First, different characters may not necessarily share the same vocabulary space, and classical NLP algorithms can hardly consume this difference. Take court debate as an example (see Fig. \ref{fig:case}). The judge can be more responsible for investigating the facts and reading the court rules while the other litigants answer the questions from the judge. Moreover, with opposite position, plaintiff and defendant's attitudes, sentiments and descriptions to the same topic can be quite different. The second barrier comes from the interactive nature of the dialogue where single utterance, without dialogue-context, barely contains enough semantics. For instance, as Fig. \ref{fig:case} shows, to accurately represent answer from the defendant, judge's question can be critical and necessary. Thus modeling the relationship among adjacent utterances across various parties is essential for dialogue context representation learning. In addition, given the colloquial dialogue content, the external knowledge-base can play a nontrivial role for context representation learning, e.g., the related law articles and legal knowledge graph can provide important auxiliary semantic information to the target trial debate. 


Motivated by such observations, in this paper, we explore dialogue context representation learning through four unsupervised pretraining tasks where the training objectives are given naturally according to the nature of the utterance and the structure of the multi-role conversation. Meanwhile, to address information implicitly, the proposed method enables the dialogue pretraining via joint learning from external knowledge resource(s). To be specific, our proposed tasks of word prediction, role prediction and utterance generation aim at learning high quality representation by randomly masking and recovering the unit component of the dialogue. The auxiliary task of reference prediction is designed for dialogue domain knowledge contextualization.

Our pretraining mechanism is fine-tuned and evaluated on three different dialogue datasets - Court Debate Dataset, Customer Service Dataset and English Meeting Dataset, with two types of downstream tasks - classification and text generation. In the experiments, we mainly testify the significance of each component in the proposed pretraining mechanism with a delicately designed encoder over court debate corpora in legal domain due to its complexity and high dependence on domain knowledge. To verify the generalizability of the proposed pretraining framework, we conduct evaluation on all downstream tasks over the other two datasets. Result shows that the proposed pretraining mechanism can significantly enhance the performance of all downstream tasks without discrimination to different encoders. Furthermore, we provide a new method to integrate multiple resources during pretraining to enrich the dialogue context.

To the best of our knowledge, this work is the pioneer investigation of multi-role dialogue pretraining with multi-tasks and multi-sources. The contributions of this study are as follows: (1) we delicately define four unsupervised pretraining objectives by masking and recovering the unit component in the dialogue context, and all pretraining tasks show positive effects on improving the testing downstream tasks; (2) we propose an innovative and effective pretraining strategy which can be generalized for different domains and different encoders; (3) In the case of small corpus, which is common for dialogue tasks, the proposed pretraining mechanism can be especially effective for quick convergence; (4) To motivate other scholars to investigate this novel but important problem, we make the experiment dataset publicly available.

\section{Problem Formulation}
Let $D=\{U_1,U_2,...,U_L\}$ denote an arbitrary dialogue, containing $L$ utterances where each utterance $U_i$ is composed of a sequence of $l$ words (namely sentence) $S_i=\{w_{i1},w_{i2},...,w_{il}\}$ and the associated role (of the speaker) $r_i$. As an optional input, for some datasets, dialogue $D$ can associate with a set of $M$ cited references $F=\{f_1,f_2,...f_M\}$ (e.g., the name of laws in the legal domain cited by the judge during the trial as shown in Fig. \ref{fig:case}). In our pretraining schema, we aim at learning high-quality representation of a dialogue by \textit{masking} and then \textit{recovering} its unit components, i.e., word, role, sentence and reference, as well as leveraging multiple resources, e.g., laws in legal domain for trial dialogue. To be clarified, the definition of important notations in the following sections are illustrated as follows:
\begin{itemize}
\setlength{\itemsep}{0pt}
\setlength{\parsep}{0pt}
\setlength{\parskip}{0pt}
    \item $D$: a debate dialogue containing $L$ utterances;
    \item $U_i$: the $i_{th}$ utterance in $D$;
    \item $r_i$: the role of the speaker in $U_i$ (i.e. judge, plaintiff, defendant and witness);
    \item $S_i$: the text content of $U_i$;
    \item $w_{ij}$: the $j_{th}$ word in $S_i$;
    \item $F$: the set of cited references in dialogue $D$ (optional);
    \item $f_m$: a cited reference in $F$;
    \item $\hat{r}_i$: the predicted role of the speaker in $U_i$;
    \item $\hat{w}_{ij}$: the predicted $j_{th}$ word in $S_i$;
    \item $\hat{S}_i$: the generated text content of $U_i$;
    \item $\hat{F}=\{\hat{f}_1,\hat{f}_2,...\hat{f}_M\}$: the predicted set of cited references in dialogue $D$ (optional)
\end{itemize}
Note that \bm{$U_i$}, \bm{$r_i$}, \bm{$S_i$}, \bm{$w_{ij}$}, \bm{${f_m}$}, \bm{$\hat{r}_i$}, \bm{$\hat{S}_i$}, \bm{$\hat{w}_{ij}$}, and \bm{${\hat{f}_m}$}  represent the embedding representations of the corresponding variables in the list. 

\section{Multi-task Masking Strategy}
\begin{figure}[!t]
    \centering
    \includegraphics[width=\linewidth]{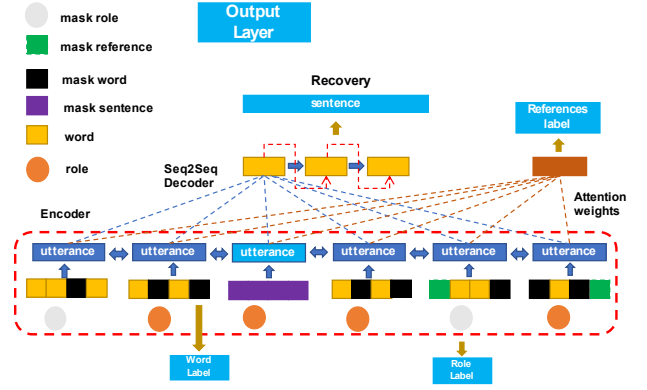}
    \caption{Concept Overview of Multi-task Masking Strategy \label{fig:model}}
    \vspace{-12pt}
\end{figure}
\subsection{Multi-Role Dialogue Encoder}
\vspace{-6pt}
In this section, we first introduce the proposed encoder for delicately representing the hierarchical information in a dialogue, and later in the experiment, we will show that the proposed pretraining mechanism can significantly enhance the performance of downstream tasks without discrimination to different encoders. 
\vspace{-12pt}
\paragraph{\makebox[1em][l]{$\,\bullet$}Utterance Layer}
In the utterance layer, we utilize a bidirectional-LSTM to encoder the semantics of the utterance while maintaining its syntactics. To involve the role information into the utterance, we concatenate the role information with each word in the sentence, which is able to project the same word into different dimensional spaces w.r.t. the target role. We hypothesize that the same word may need differentiate when different speakers use it.  
\setlength{\abovedisplayskip}{4pt}
\setlength{\belowdisplayskip}{4pt}
\begin{align*}
 \mathbf{h_{ij}} = \left [ \overrightarrow{\mathbf{LSTM^U}}(\mathbf{e_{ij}}); \overleftarrow{\mathbf{LSTM^U}}(\mathbf{e_{ij}})\right], j \in \left[1, l \right]
\end{align*}
where $\mathbf{e_{ij}} = [\mathbf{w_{ij}}; \mathbf{r_i}]$. To strengthen the relevance between words in an utterance, we employ the attention mechanism to obtain $\mathbf U_i$, which can be interpreted as a local representation of an utterance:
\setlength{\abovedisplayskip}{4pt}
\setlength{\belowdisplayskip}{4pt}
\begin{align*}
 & \mathbf{U_{i}} = \sum_{j=1}^{l} \alpha_j^u \mathbf{h_{ij}}\\ 
 & \alpha_j^u =   \frac{\exp(\mathbf{Q^uh_{ij}})}{\sum_{{j}'=1}^{l}\exp(\mathbf{Q^uh_{i{j}'}})}
\end{align*}
where $\mathbf{Q^u}$ are learnable parameters.
\vspace{-6pt}
\paragraph{\makebox[1em][l]{$\,\bullet$}Dialogue Layer}
To represent the global context in a dialogue, we employ another bidirectional-LSTM to encode the dependency between utterances to obtain a global representation of an utterance, denoted as $\mathbf{\overline{U}_i}$:
\begin{align*}
 &\mathbf{\overline{U}_i} = \left [\overrightarrow{\mathbf{LSTM^D}}(\mathbf{U_{i}}); \overleftarrow{\mathbf{LSTM^D}}(\mathbf{U_{i}}) \right], i \in \left[ 1, L\right] \\
 &\mathbf{\overline{U}} =\left\{\mathbf{\overline{U}_1},\mathbf{\overline{U}_2},..., \mathbf{\overline{U}_L}\right\} \in \mathbf{R}^{L\times2dim_h}
\end{align*}
Then, we feed $\mathbf{\overline{U}}$ to a N-layer Transformer-Block\cite{vaswani2017attention} to suppress the long-term dependency for long dialogue, and finally obtain a dialogue representation $\mathbf{\widetilde{U}} = \mathbf{Transformer^{N}}(\mathbf{\overline{U}})$, which will be used in the following pretraining tasks as global dialogue context representation. 
\paragraph{\makebox[1em][l]{$\,\bullet$}Knowledge Enhance Layer}
For some dataset associated with domain knowledge (e.g., in the court debate dataset, the dialogue context can highly rely on the legal domain knowledge, e.g., laws and logic), we propose a Knowledge Enhance Layer to enable external knowledge/resource integration into the utterance representation. At the Knowledge Enhance Layer, the representation of dialogue is enhanced by quoting the content of articles of law, for instance, in the court debate scenario (the masked green parts as shown in Fig. \ref{fig:case}). In order to enhance the dialogue representation learning performance from the legal knowledge viewpoint, the proposed dialogue representation is able to learn the vital information from the articles of law in the case context through the attention mechanism. Given a set of cited references $F$ (e.g., all the laws for court debate), we use $C=\{w_{1}^{F},w_{2}^{F},...,w^{F}_t\}$ to represent all the content in the references and the corresponding embedding representations is $\mathbf{C} =\{\mathbf{w_{1}^{F}},\mathbf{w_{2}^{F}},...,\mathbf{w^{F}_t}\}$. The word embeddings are shared with the words in dialogues. We employ Bi-LSTM to encode the context semantics of $C$ and apply attention mechanism to address the relevance between utterance and the reference:
\setlength{\abovedisplayskip}{4pt}
\setlength{\belowdisplayskip}{4pt}
\begin{align*}
\mathbf{h_{n}^{F}} = &\left [\overrightarrow{\mathbf{LSTM^F}}(\mathbf{w_{n}^F}); \overleftarrow{\mathbf{LSTM^F}}(\mathbf{w^F_{n}}) \right], n \in \left[ 1, t\right] \\
& \mathbf{\overline{C}_{i}} = \sum_{n=1}^{t} \alpha_{in}^c \mathbf{h_{n}^{F}}\\ 
 & \alpha_{in}^c =   \frac{\exp(\mathbf{\widetilde{U}_{i}Q^{c}h^F_{n}})}{\sum_{{n}'=1}^{t}\exp(\mathbf{\widetilde{U}_{i}Q^{c}h^F_{{n}'}})}
\end{align*}
where $\mathbf{w_{t}^F}$ is the $t_{th}$ word in the content of references (e.g., laws) cited in the current dialogue. $\mathbf{Q^c}$ are learnable parameters.

\subsection{Pretraining with Multi-task Masking Strategy}
To host heterogeneous information/knowledge in a dialogue pretraining, in this section, we propose a Multi-task Masking Strategy, which is able to optimize the dialogue representation in terms of four different prediction tasks. The concept overview of the proposed strategy is depicted in Fig. \ref{fig:model}.
\paragraph{\makebox[1em][l]{$\,\bullet$}Reference Prediction (F.P.)}
Reference prediction is a multi-label classification task for an entire dialogue, which aims at recovering the masked references in a given dialogue. In the experiment, we conduct this masking strategy for court debate dataset where we mask the article names (e.g., Article 8 (4) of the Contract Law) by mining top frequent article names existing in judgment documents. The predicted representation of the references is $\mathbf{\hat{f}}= g(\mathbf{V^f} f_p(\mathbf{\widetilde{U}}) + \mathbf{b^f})$,where $g(\ast )$ is a non-linear activation function, $f_p$ is a pooling function and $\mathbf{Q^f}$,$\mathbf{V^f}$,$\mathbf{b^f}$ are learnable parameters. Finally, we pass $\mathbf{\hat{f} }$ to a fully connected layer and then to a sigmoid function layer for reference prediction. $y_k^{f}$ denotes the ground truth label and  $p_k^{f}$ is its predicted label. The binary cross-entropy loss function is applied:
\setlength{\abovedisplayskip}{4pt}
\setlength{\belowdisplayskip}{-12pt}
 \begin{align*}
 & f_p(\mathbf{\widetilde{U}})  = \sum_{j=1}^{L} \alpha_j^f \mathbf{\widetilde{U}_{j}}\\ 
 & \alpha_j^f =   \frac{\exp(\mathbf{Q^f\widetilde{U}_{j}})}{\sum_{{i}'=1}^{L}\exp(\mathbf{Q^f\widetilde{U}_{{i}'}})}
\end{align*}
 $$L^{f} = \sum_{k=1}^{F}  \left[  y_k^f\log(p_k^f)  + (1 - y_k^f)\log(1 - p_k^f)\right]$$

\paragraph{\makebox[1em][l]{$\,\bullet$}Word Prediction (W.P.)}
Word prediction is a multi-class classification task. $Z$ denotes the set of all the masked words in dialogue $D$\footnote{Base on the prior experience in \cite{devlin2018bert}, we randomly mask $15\%$ words for each sentence}. For arbitrary sentence $S_i$ and arbitrary masked word $w_{ij}$ in $S_i$, the predicted representation of the masked word $w_{ij}$ is $\mathbf{\hat{w}_{ij} }= g(\mathbf{V^w}\left [\mathbf{h_{ij}}; \mathbf{\widetilde{U}_i}; \mathbf{\overline{C}_i}  \right] +  \mathbf{b^w})$, where $\mathbf{V^w}$ and $\mathbf{b^w}$ are learnable parameters. $\mathbf{h_{ij}}$ is to fetch local contextual information from $S_i$ and $\mathbf{\widetilde{U}_i}$ is used to enhance the global contextual information from dialogue $D$, and $\mathbf{\overline{C}_i}$ helps to involve the external knowledge. Finally, we pass $\mathbf{\hat{w}_{ij} }$ to a fully connected layer and then to a softmax function for word prediction. $y_{zk}^{w}$ denotes the ground truth word $z\in Z$ and $p_{zk}^{w}$ is the predicted word. The cross-entropy loss function is:
\setlength{\abovedisplayskip}{4pt}
\setlength{\belowdisplayskip}{0pt}
\begin{align*}
L^{w} = -\sum_{z\in Z}\sum_{k=1}^{N^{w}} y_{zk}^{w}\log(p_{zk}^{w}))
\end{align*}
\vspace{-6pt}

\paragraph{\makebox[1em][l]{$\,\bullet$}Role Prediction (R.P.)}
Role prediction is also a multi-class classification task. $G$ are the set of all the masked roles in a dialogue\footnote{In the experiment, we randomly mask the roles of $15\%$ utterances for each dialogue.}. For arbitrary utterance $U_i$ and corresponding masked role $r_i$, the predicted representation of the masked role is $\mathbf{\hat{r}_{i}}= g(\mathbf{V^r} \left[\mathbf{\widetilde{U}_i}; \mathbf{\overline{C}_i}\right] + \mathbf{b^r})$, where $\mathbf{V^r}$ and $\mathbf{b^r}$ are learnable parameters. Finally, we pass $\mathbf{\hat{r}_{i}}$ to a fully connected layer and to a softmax function for role prediction. $y_{gk}^{r}$ denotes the ground truth role label and  $p_{gk}^{r}$ is the predicted role label. The cross-entropy loss function is:
\setlength{\abovedisplayskip}{4pt}
\setlength{\belowdisplayskip}{4pt}
\begin{align*}
 L^{r} = -\sum_{g\in G}\sum_{k=1}^{N^{r}} y_{gk}^r\log(p_{gk}^r)
\end{align*}

\paragraph{\makebox[1em][l]{$\,\bullet$}Sentence Generation (S.G.)}
Sentence Generation is an NLG task. We utilize encoder-decoder framework with attention mechanism \cite{luong2015effective} for pretraining. We use LSTM cell as basic decoder cell, and $\{[\mathbf{\widetilde{U}_1}; \mathbf{\overline{C}_1}],[\mathbf{\widetilde{U}_2}; \mathbf{\overline{C}_2}],...,[\mathbf{\widetilde{U}_L}; \mathbf{\overline{C}_L}]\}$ is the encoder representation for masked sentence $S_{i}$ in dialogue $D$\footnote{
In the experiment, we randomly sample one sentence from dialogue $D$ according to the prior experience of a similar task conducted in \cite{mehri2019pretraining}.}. The loss function is:
\setlength{\abovedisplayskip}{4pt}
\setlength{\belowdisplayskip}{4pt}
\begin{align*}
 &L^{s}  = -\log P(S_i\mid D)\\ 
 & \quad \  \, =  - \sum_{k=1}^{l}\log P(w_{ik} \mid w_{i1:k-1}, D)
\end{align*}

The final loss function of the four pretraining objectives is shown as below: $$L^{total} = L^{f} + L^{w} + L^{r} + L^{s}$$ which encapsulates various kinds of semantics/knowledge for dialogue pretraining via multi-task masking.

\section{Evaluated Downstream Tasks}

\begin{table}[!t]
\caption{Statistics of Three Corpus for Pretraining and Downstream Tasks. Note that the \#length denotes on average the number of utterances in a dialogue of each corpus.}
\label{Tab:dataset}
\small
\centering
\resizebox{\columnwidth}{!}{\begin{tabular}{c|ccc|cc}
\hline
       & \multicolumn{3}{c|}{Pretraining} & \multicolumn{2}{c}{Downstream Tasks} \\ 
Corpus & \#utterance     & \#dialogue   &\#length & \#utterance   & \#dialogue    \\ \hline
\texttt{CDD}    & 20M              & 340K  &59       & 1.6M          & 6,129           \\
\texttt{CSD}    & 70M              & 5M     & 14     & 130K          & 3,463           \\
\texttt{EMD}    & 1M               & 32K    & 31      & 73K         & 7,824         
\\ \hline
\end{tabular}}
\end{table}

\begin{table}[!t]
\caption{Pretraining Results over Three Corpus.}
\label{Tab:pretraining}
\centering
\begin{tabular}{c|c|c|c|c}
\hline
Corpus & W.P./acc.    & R.P./acc.   & S.G./bleu    & F.P./acc.    \\ \hline
\texttt{CDD}    & 77.88 & 84.54 & 37.30 & 96.34 \\ 
\texttt{CSD}    & 53.82 & 94.96 & 18.08 & -     \\ 
\texttt{EMD}    & 62.63 & -     & 57.74 & -    \\ \hline
\end{tabular}
\end{table}
To validate the performance and generality of the proposed pretraining mechanism, in this section, we evaluate two types of downstream tasks, \textit{classification} and \textit{summarization}, over three open multi-role dialogue datasets.

\subsection{Datasets}
\paragraph{\makebox[1em][l]{$\,\bullet$}Court Debate Dataset (\texttt{CDD})}
\texttt{CDD} corpus contains over $340$K court debate records of civil private loan disputes cases. The court record is a multi-role debate dialogue associating four roles, i.e., judge, plaintiff, defendant and witness. According to the statistics as shown in Table \ref{Tab:dataset}, court debate appears to have longer conversations, on average, containing $59$ utterances in a dialogue. We release all the experiment data to motivate other scholars to further investigate this problem\footnote{\url{https://github.com/wangtianyiftd/dialogue_pretrain}}.  

\paragraph{\makebox[1em][l]{$\,\bullet$}Customer Service Dialogue (\texttt{CSD})}
\texttt{CSD} corpus\footnote{https://sites.google.com/view/nlp-ssa} is collected from the customer service center of a top E-commerce platform, which contains over $5$ million customer service records between two roles (customer and agent) related to two product categories namely \textit{Clothes} and \textit{Makeup}. The statistics of the dataset is shown in Table \ref{Tab:dataset}.   
\vspace{-12pt}
\paragraph{\makebox[1em][l]{$\,\bullet$}English Meeting Dataset (\texttt{EMD})}
\texttt{EMD} corpus is a combined dataset\footnote{For pre-training considerations, relatively large amount of data is required. Thus we combine the four open datasets for pretraining.} consisting of four open English meeting corpus: AMI-Corpus\cite{goo2018abstractive}, Switchboard-Corpus\cite{jurafsky2000speech}, MRDA-Corpus\cite{shriberg2004icsi} and bAbI-Tasks-Corpus\footnote{\url{https://github.com/NathanDuran/bAbI-Tasks-Corpus}}. Among the above four corpus, AMI-Corpus includes manually annotated act labels and summaries for the meeting conversations, thus we use such annotated data in AMI-Corpus for downstream tasks. Comparing with the other two datasets, EMD can be much smaller. We use it to validate our hypothesis that the proposed pretraining mechanism can be also efficient for small dialogue corpus.

\subsection{Classification}
Judicial Fact Recognition (\textbf{JFR}) is a multi-class classification task, specifically for the court debate corpus in legal domain. The identified judicial facts are the key factors for the judge to analyze and make decision of the target case, thus the objective of this task is to assign each utterance in the court debate to either one of the judicial facts\footnote{$10$ fact labels are used in this task: \textit{principal dispute}, \textit{guarantee liability}, \textit{couple debt}, \textit{interest dispute}, \textit{litigation statute dispute}, \textit{fraud loan}, \textit{liquidated damages}, \textit{involving criminal proceedings}, \textit{false lawsuit} and \textit{creditor qualification dispute}.} (or the category of \textit{Noise})\footnote{Statistically, $71.2\%$ of utterances in the experiment data are regarded as noises, i.e., independent to the judicial elements.}, to represent the correlation between the utterance and the essential facts. This task mainly evaluates the strength of the pretraining framework on representing the semantics of the complex multi-role debate context as well as differentiating the informative context from the noisy content.

Dialogue Act Recognition (\textbf{DAR}) is also a multi-class classification task conducted over the CSD and EMD corpus respectively. The labels in CSD corpus characterize the actions of both customer and staff, i.e., customer side's acts - \textit{advisory},  \textit{
request operation}, and staff side's acts - \textit{
courtesy reply}, \textit{answer customer's question}, and in total $14$ labels are involved. Similar in EMD corpus, each utterance is assigned with an act label out of $15$ possible labels. 

\subsection{Text Generation}
Controversy Focus Generation (\textbf{CFG}) is an abstractive summarization task for the court debate corpus. During the civil trial, the presiding judge summarizes the essential controversy focuses\footnote{Here shows two examples of the summarized controversy focuses: \textit{Is the loan relationship between the plaintiff and defendant established? Did the plaintiff fulfill he obligation to lend the money?}} according to the plaintiff's complaint\footnote{The plaintiff's claiming legal rights against another. \url{https://dictionary.law.com/Default.aspx?selected=261}} and the defendant's answer\footnote{The defendant's pleading to respond to a complaint in a lawsuit. \url{https://dictionary.law.com/Default.aspx?selected=2407}}. Later, the parties from different camps (plaintiff, defendant, witness etc.) debate on court with each other based on the controversy focuses summarized by the presiding judge. This task is challenging in that the construction of abstractive summarization between debated dialogue requires high-quality of global context representation of the entire dialogue which captures the correlation among the utterances by different characters. Thus the pretraining process tends to be significant for initializing parameters of hidden states for the decoders especially when it comes with limited size of training data. 

Dialogue Summarization (\textbf{DS}) aims at generating a summary for a given dialogue and this task is conducted over EMD corpus. Compared to the text generation task for court debate, the annotated summary in EMD corpus is much shorter and is mainly comprised of key phrases instead of long sentences which describes the topic/intent of a given meeting dialogue\footnote{Here shows several examples of the annotated meeting summaries: \textit{evaluation of project process}, \textit{possible issues with project goals}, \textit{closing}, \textit{discussion}, \textit{marketing strategy}.}.

\subsection{Initialization for Downstream Tasks Training}
In the training phase of downstream tasks, for the classification, we use the pretrained dialogue representation (because the dialogue representation has gathered word, sentence and role representation) to initialize the dialogue representation of the classification tasks. The parameters of the decoder part in classification model are randomly initialized (including ones in softmax layer, full connection layer); As for the generation task, we use the pretrained dialogue representation and the parameters of the decoding part (LSTM cell and attention) of the Sentence Generation task (one of the four pretraining tasks), because the decoder structure of the downstream text generation task is the same as that of the pretraining task.
\section{Experimental Settings}

\begin{table*}[!t]
\caption{Downstream Task - Classification Results with Different Encoders over Three Corpus. $\dagger$ at ``pretrain'' rows indicates statistically significant difference from the corresponding value of ``vanilla'' model ($p<0.01$).}
\label{Tab:classification}
\centering
\begin{tabular}{c|c|c|c|c|c|c}
\hline
\multirow{2}{*}{Model}    & \multicolumn{2}{c|}{\texttt{CDD}: JFR} & \multicolumn{2}{c|}{\texttt{CSD}: DAR} & \multicolumn{2}{c}{\texttt{EMD}: DAR} \\
          &\multicolumn{1}{c}{micro F$_1$}   & \multicolumn{1}{c|}{macro F$_1$}   & \multicolumn{1}{c}{micro F$_1$}   & \multicolumn{1}{c|}{macro F$_1$}   & \multicolumn{1}{c}{micro F$_1$}   & \multicolumn{1}{c}{macro F$_1$}   \\ \hline
HBLSTM-CRF(vanilla)  & 81.60      & 34.14      & 84.96      & 76.03      & 66.04      & 51.53      \\
HBLSTM-CRF(pretrain) & \textbf{81.68}$\dagger$      & \textbf{39.13}$\dagger$      & \textbf{85.34}$\dagger$      & \textbf{77.36}$\dagger$      & \textbf{66.22}     & \textbf{51.94}      \\ 
CRF-ASN(vanilla)     & 80.90      & 31.15      & 83.55      & 73.75      & 64.13      & 44.12      \\
CRF-ASN(pretrain)    & \textbf{81.55}$\dagger$      & \textbf{38.22}$\dagger$      & \textbf{83.92}$\dagger$     & \textbf{74.93}$\dagger$      & \textbf{64.58}$\dagger$      & \textbf{49.55}$\dagger$      \\ \
Our model(vanilla)   & 81.73      & 39.61      & 85.34      & 76.83      & 66.83      & 52.78      \\
Our model(pretrain)  & \textbf{82.06}$\dagger$      & \textbf{45.45}$\dagger$      & \textbf{85.80}$\dagger$      & \textbf{78.28}$\dagger$      & \textbf{67.34}$\dagger$      &\textbf{53.69}$\dagger$    \\ \hline 
\end{tabular}
\end{table*}

\begin{table*}[!t]
\caption{Downstream Task - Summarization Results with Different Encoders over Two Corpus. $\star$ at ``pretrain'' rows indicates statistically significant difference from the corresponding value of ``vanilla'' model ($p<0.001$).}
\label{Tab:summarization}
\centering
\resizebox{\linewidth}{!}{\begin{tabular}{c|ccccc|ccccc}
\hline
\multirow{2}{*}{Model} & \multicolumn{5}{c|}{\texttt{CDD}: CFG}                  & \multicolumn{5}{c}{\texttt{EMD}: DS}                   \\
                       & rouge-1 & rouge-2 & rouge-3 & rouge-L & bleu4 & rouge-1 & rouge-2 & rouge-3 & rouge-L & bleu4 \\ \hline
DAHS2S(vanilla)        & 26.83   & 7.27   & 4.15   & 22.21   & 5.27  & 35.43   & 28.84   & 25.95   & 34.33   & 20.69 \\
DAHS2S(pretrain)       & \textbf{34.94}$\star$   & \textbf{12.98}$\star$   & \textbf{7.18}$\star$   & \textbf{29.51}$\star$   & \textbf{8.06}$\star$  & \textbf{40.68}$\star$   & \textbf{32.11}$\star$   & \textbf{27.00}$\star$   & \textbf{39.60}$\star$   & \textbf{22.26}$\star$ \\
Our model(vanilla)     & 22.55   & 3.99   & 1.66   & 18.95   & 2.94  & 31.00   & 26.40   & 25.23   & 29.83   & 14.87 \\
Our model(pretrain)    & \textbf{36.55}$\star$   & \textbf{13.54}$\star$   & \textbf{7.48}$\star$   & \textbf{30.84}$\star$   & \textbf{8.59}$\star$  & \textbf{41.39}$\star$   & \textbf{34.18}$\star$   & \textbf{29.64}$\star$   & \textbf{40.34}$\star$   & \textbf{23.20}$\star$ \\ \hline
\end{tabular}}
\end{table*}

\subsection{Tested Encoders}
In the experiment, for each downstream task, we perform pretraining with several state-of-the-art encoders as well as our proposed encoder in this paper. This experiment setting ensures a universality validation of the proposed pretraining mechanism over a variety of encoding strategies.

For the \textit{classification} tasks, besides our proposed encoder, we also select two models - \textbf{HBLSTM-CRF}\cite{kumar2018dialogue} and \textbf{CRF-ASN}\cite{chen2018dialogue} - as encoders for both pretraining and downstreaming stages. The two models have leading performance on MRDA-Corpus and Switchboard-Corpus as shown on the Leaderboard\footnote{\url{http://nlpprogress.com/english/dialogue.html}}. 

As for the \textit{text generation} tasks in \texttt{CDD} and \texttt{EMD} corpus, except for using our own encoder, we use Discourse Aware Hierarchical Sequence-to-Sequence model (\textbf{DAHS2S})\cite{cohan2018discourse} employed in \cite{goo2018abstractive} as the other encoder for abstractive summarization.

\subsection{Evaluation Metrics}
For \textit{classification} task, we evaluate the performance of each model based on two popular classification metrics: micro F1 and macro F1 scores.
To automatically assess the quality of \textit{generated text}, we used ROUGE \cite{lin2003automatic} and BLEU\cite{papineni2002bleu} scores to compare different models. We report ROUGE-1 and ROUGE-2 as the means of assessing informativeness and ROUGE-L as well as BLEU-4 for assessing fluency.

\subsection{Hyper-Parameter Selection}
In our experiments, we optimize the tested models using Adam Optimization\cite{kingma2014adam} with learning rate of $5e$-$4$. The dimensions of word embedding and role embedding are $300$ and $100$ respectively. The size of hidden layers are all set to $256$. We use $2$ layer Transformer-Block, where feed-forward filter size is $1024$, and the number of heads equals to $4$.

\section{Results and Discussion}
\subsection{Overall Performance}

To evaluate the performance of the proposed pretraining model, we export the results of pretraining tasks as well as the improved performance on downstream tasks over three different datasets. Table \ref{Tab:pretraining} shows the performance of the proposed pretraining tasks on all datasets\footnote{Note that there is no reference used in \texttt{CSD} and \texttt{EMD} corpus and \texttt{EMD} corpus contains no character information neither, so the corresponding pretraining tasks are omitted for the two corpus.} which indicates how effective the proposed pretraining mechanism on recovering the information in the dialogue. According to the pretraining scores, we can also observe the complexity of the corresponding corpus. For instance, The performance of word prediction and sentence generation tasks in \texttt{CSD} corpus is worse than that in \texttt{CDD} corpus due to the word diversity in customer service for coping with a variety of disputes, however in relatively close legal domain, the words of different roles especially of the judges during trial remain similar across different cases. As for the role prediction, in customer service, usually only two characters are involved while in court debate it is common to have multiple roles therefore it might be the reason why the task of role prediction in \texttt{CDD} is lower than that in \texttt{CSD}.

\begin{figure*}[!t]
\centering
\begin{subfigure}{.5\textwidth}
  \centering
  \includegraphics[width=0.9\linewidth]{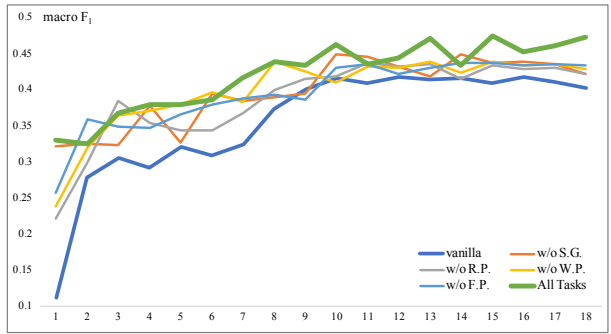}
  \caption{Task JFR (macro F$_1$ score)}
  \label{fig:convergence-JFR}
\end{subfigure}%
\begin{subfigure}{.5\textwidth}
  \centering
  \includegraphics[width=0.9\linewidth]{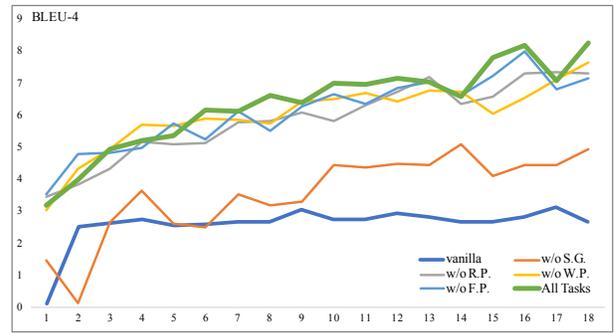}
  \caption{Task CFG (BLEU-4 score)}
  \label{fig:convergence-CFG}
\end{subfigure}
\caption{The performance of two downstream tasks with different pretraining objectives.}
\vspace{-12pt}
\label{fig:convergence}
\end{figure*}

Table \ref{Tab:classification} and \ref{Tab:summarization} demonstrate the performance on two downstream tasks respectively over three datasets where the notation ``vanilla'' represents the randomly initialized encoders used for the downstream tasks while ``pretrain'' denotes the pretrained encoders for the downstream tasks.
\paragraph{\makebox[1em][l]{$\,\bullet$}Classification} 
Table \ref{Tab:classification} aggregates the classification results on all three datasets. In general, we can observe that, for all tested encoders, the pretraining process results are significantly superior than baselines over all datasets under almost all metrics, especially for macro F$_1$ score. Statistically, on average, pretraining under different encoders achieves $17.4\%$, $1.4\%$ and $5\%$ increase in macro F$_1$ score over \texttt{CDD}, \texttt{CSD} and \texttt{EMD} corpus respectively, which implies such pretraining can be very helpful to alleviate the problem of unbalanced/bias data where there are some categories with rather small/large training data (e.g., in \texttt{CDD} corpus, the category of \textit{Noise} takes up more than $70\%$ of utterances.). Moreover, our proposed encoder in this paper outperforms the state-of-the-art encoders for the tested corpus. 
\paragraph{\makebox[1em][l]{$\,\bullet$}Text Generation}
Table \ref{Tab:summarization} depicts the text generation results on \texttt{CDD} and \texttt{EMD} corpus respectively. Similar to classification task, the pretraining conducted under both encoders shows positive effects on all the evaluation metrics. An interesting finding is that the proposed model achieves limited performance in ``vanilla'' setting but after pretraining it quickly surpassed the method DAHS2S in all metrics. In addition, as we aforementioned, the \textbf{CFG} task is much challenging compared to the \textbf{DS} task due to the relatively long text (i.e., controversy focus) needs to be generated via \texttt{CDD} corpus. In such difficult case (\texttt{CDD} corpus), the proposed pretraining method performs beyond expectation. While it successfully estimates the comprehensive dialogue context representation, comparing with the baselines, Bleu-4 scores increased by $53\%$ and $192\%$. As for the relatively small dataset, \texttt{EMD} corpus, the pretraining method brings about $7.6\%$ and $56\%$ increase for the two tested encoders.

\subsection{Ablation Test}
To assess the contribution of different components in the proposed method, we conduct ablation tests for both classification (see Table \ref{Tab:ablation-classification}) and text generation (see Table \ref{Tab:ablation-summarization}) tasks on \texttt{CDD} corpus\footnote{Only \texttt{CDD} corpus enables testing on all four pretraining objectives (see Table \ref{Tab:pretraining}).}. To prove the generalizability of the proposed pretraining schema on different encoders, for all tested encoders, the same ablation test is conducted by removing each pretraining objective. 

Table \ref{Tab:ablation-classification} reports the F$_1$ scores of \textbf{JFR} task, for each encoder, when training on all objectives and when training on all objectives except the particular one. As Table \ref{Tab:ablation-classification} shows, all the model components contribute positively to the results. To be specific, the pretraining task of word prediction has largest impact on the method HBLSTM-CRF. Their removal causes $1.8\%$ relative increase in error (RIE) for macro F$_1$ scores, while the task of role prediction has biggest impact on the model ASN-CRF ($9.3\%$ RIE for macro F$_1$ score). As for our encoder, reference and word prediction show greatest impact on the performance. In general, we notice that, for classification, the three prediction tasks affect the model effect to varying degrees.

As for \textbf{CFG} task, the findings are quite different as suggested in Table \ref{Tab:ablation-summarization}. Since \textbf{CFG} is a text generation task, we can observe that the pretraining task, sentence generation, tends to have largest impact on both tested encoders evaluated by Bleu-4 score. Such observations indicate that the pretraining tasks have strong impact on the downstream tasks in similar types.    

\begin{table}[!t]
\caption{Ablation Test on \textbf{JFR} Tasks with Different Encoders over Court Debate Dataset}
\label{Tab:ablation-classification}
\centering
\resizebox{\columnwidth}{!}{\begin{tabular}{c|cc|cc|cc}
\hline
\multirow{2}{*}{Method} & \multicolumn{2}{c|}{HBLSTM-CRF} & \multicolumn{2}{c|}{ASN-CRF} & \multicolumn{2}{c}{Our Model} \\
                        & miF$_1$       & maF$_1$      & miF$_1$     & maF$_1$     & miF$_1$      & maF$_1$      \\ \hline
All                & \textbf{81.68}          & \textbf{39.13}         & \textbf{81.55}        & \textbf{38.22}        & \textbf{82.06}         & \textbf{45.45}         \\
w/o W.P.                & 81.63          & 38.03         & 81.15        & 34.17        & 81.77         & 41.66        \\
w/o R.P.                & 81.64          & 38.38         & 81.06        & 32.46        & 82.01         & 40.78         \\
w/o S.G.                & 81.66          & 38.13         & 81.27        & 32.93        & 81.97         & 43.81         \\
w/o F.P.                & 81.65          & 38.41         & 81.47       & 35.07        & 81.73         & 41.71        \\ \hline
\end{tabular}}
\vspace{-6pt}
\end{table}
\vspace{-6pt}

\begin{table}[!htbp]
\caption{Ablation Test on \textbf{CFG} Tasks with Different Encoders over Court Debate Dataset}
\label{Tab:ablation-summarization}
\centering
\resizebox{\columnwidth}{!}{\begin{tabular}{cccccc}
\hline
Method               & rouge-1 & rouge-2 & rouge-3 & rouge-L & bleu4 \\ \hline
DAHS2S        & 34.94   & \textbf{12.98}  & \textbf{7.18}  & 29.51   & \textbf{8.06}  \\
w/o W.P. & 34.65   & 11.98   & 6.64   & 28.75   & 7.92  \\
w/o R.P. & 34.26   & 12.13   & 6.39   & 28.54   & 7.42  \\
w/o S.G.    & 30.93   & 9.64  & 5.38   & 25.74   & 6.40  \\
w/o F.P.  & \textbf{35.54}   & 12.69   & 6.82   & \textbf{29.77}   & 7.92  \\ \hline
Our Model     & \textbf{36.55}   & \textbf{13.54}   & \textbf{7.48}   & \textbf{30.84}   & \textbf{8.59}  \\
w/o W.P. & 35.79   & 13.13   & 7.10   & 29.96  & 8.09  \\
w/o R.P. & 35.58   & 12.59   & 7.10   & 30.06   & 8.22  \\
w/o S.G.    & 30.38   & 8.59   & 4.25   & 24.85   & 5.58  \\
w/o F.P.  & 36.49   & 13.13   & 7.21   & 30.64   & 8.31 \\ \hline
\end{tabular}}
\vspace{-6pt}
\end{table}
\vspace{-6pt}
\subsection{Convergence Analysis}

To further validate the performance of the proposed pretraining model, we conduct experiments to monitor the impact of pretraining on the convergence of all downstream tasks. In the experiment, we employ the proposed model as encoder and evaluate the performance of two downstream tasks with pretraining on all objectives and on all objectives except the particular one at each epoch. Fig. \ref{fig:convergence-JFR} and \ref{fig:convergence-CFG} depict the results of \textbf{JFR} and \textbf{CFG} tasks respectively. 

As shown in Fig.\ref{fig:convergence-JFR} and \ref{fig:convergence-CFG}, we can observe that the performance of the model with pretraining on all objectives is significantly superior than the ``vanilla'' one from the initial epoch which indicates the advantage of learning with pretrained parameters instead of random initialization. Compared the pretrained ``all tasks'' model with the models removing a particular task, the former performs more stably and almost always outperforms the latter ones.

\section{Related Work}
\subsection{Unsupervised Pretraining in NLP}
Unsupervised pretraining for natural language becomes popular and widely adopted in many NLP tasks due to the nature that labeled data for specific learning tasks can be highly scarce and expensive. Due to such motivation, the earliest approaches used unlabeled data to compute word-level or phrase-level embeddings\cite{collobert2011natural,mikolov2013distributed,mikolov2013efficient,pennington2014glove}, which were later used as atom features in a supervised model for the further specific learning tasks. Although the pretrained word/phrase-level embeddings could improve the performance on various tasks, such approaches can only capture the atom-level information regardless of the higher-level semantics and syntactics. Recent research work have focused on learning sentence-level and document-level representations which are learned from unlabeled corpus \cite{radford2018improving,peters2018deep,chang2019language}. 

Compared to the sentence or document-level representation learning, dialogue representation learning can be more complex due to its hierarchical structures as well as its heterogeneous information resources. In this work, we address the difficulty of such challenges and propose a masking strategy for pretraining in multi-task schema.

\subsection{Dialogue Representation Learning}
Recent research work has focused on proving the effectiveness of hierarchical modeling in dialogue scenarios \cite{weston2016dialog}. The common approach is focusing on constructing delicate encoders for representing dialogue structures. For instance, Weston\cite{weston2016dialog} employed a memory network based encoder to capture the context information in a dialogue for specific task learning. 
Although there have been plenty of research work focusing on document representation learning, pretraining methods are still in their infancy in the domain of dialogue. Mehri et al\cite{mehri2019pretraining} recently approaches to such problem by proposing pretraining objectives for dialogue context representation learning. Compared to them, we are different in several aspects: Firstly, to the best of our knowledge, we are the first to involve role information in this area, and in our framework, we are flexible to involve external resources during pretraining; Secondly, all tasks in our work are in bidirectional approach which means we can consider the context in both directions, similar to the strategy to BERT\cite{devlin2018bert}; Third, the experimental results demonstrate the generalizability of our proposed pretraining strategy over different domains and along with various types of encoders. 
\section{Conclusion}
This paper investigates the research problem of dialogue context representation learning by proposing a multi-task masking strategy to perform various types unsupervised pretraining tasks, including word prediction, role prediction, sentence generation and reference prediction. The proposed fine-tuned pretraining mechanism is comprehensively evaluated through three different dialogue datasets along with a number of downstream dialogue-mining tasks. Result shows that the proposed pretraining mechanism significantly contributes to all the downstream tasks without discrimination to different encoders.
\section{Acknowledgments}
This work is supported by National Key R\&D Program of China 
(2018YFC0830200;2018YFC0830206).
\bibliographystyle{aaai}
\bibliography{aaai}
\end{document}